\documentclass[10pt,journal,compsoc]{IEEEtran}
\usepackage{amsmath,amsfonts}
\usepackage{array}
\usepackage[caption=false,font=normalsize,labelfont=sf,textfont=sf]{subfig}
\usepackage{textcomp}
\usepackage{stfloats}
\usepackage{url}
\usepackage{verbatim}
\usepackage{graphicx}
\usepackage{cite}
\usepackage{color}
\usepackage[export]{adjustbox}
\usepackage{amsmath}
\usepackage{algorithm}
\usepackage{algpseudocode}
\begin{document}

\title{Multi-instance robust fitting for non-classical geometric models}

\author{Zongliang Zhang, Shuxiang Li, Xingwang Huang, Zongyue Wang
\thanks{Zongliang Zhang (zzl@jmu.edu.cn), Shuxiang Li, Xingwang Huang (huangxw@jmu.edu.cn) and Zongyue Wang (wangzongyue@jmu.edu.cn) are with the College of Computer Engineering, Jimei University, Xiamen, China.} 
\thanks{Corresponding author: Xingwang Huang}
}



\maketitle

\begin{abstract}
Most existing robust fitting methods are designed for classical models, such as lines, circles, and planes. In contrast, fewer methods have been developed to robustly handle non-classical models, such as spiral curves, procedural character models, and free-form surfaces. Furthermore, existing methods primarily focus on reconstructing a single instance of a non-classical model. This paper aims to reconstruct multiple instances of non-classical models from noisy data. We formulate this multi-instance fitting task as an optimization problem, which comprises an estimator and an optimizer. Specifically, we propose a novel estimator based on the model-to-data error, capable of handling outliers without a predefined error threshold. Since the proposed estimator is non-differentiable with respect to the model parameters, we employ a meta-heuristic algorithm as the optimizer to seek the global optimum. The effectiveness of our method are demonstrated through experimental results on various non-classical models. The code is available at \url{https://github.com/zhangzongliang/fitting}.     
\end{abstract}


\section{Introduction}\label{sec:introduction}

\IEEEPARstart{R}{obust} model fitting aims to find model instances that best describe a set of data points contaminated by outliers \cite{xiao2024latent}. It has broad applications in science and engineering, such as point cloud segmentation \cite{li2025weakly}, point cloud registration \cite{chen2023sc}, image alignment \cite{shapson2024petavoxel}, groundwater level analysis \cite{jasechko2024rapid}, road curve reconstruction \cite{zhang20193d}, and few-shot learning \cite{peng2025robust}.

Most existing robust fitting methods are based on the RANdom SAmple Consensus (RANSAC) algorithm \cite{fischler1981random}. Assuming that a model instance can be determined by a minimal subset of data points, RANSAC-based methods generate candidate instances by sampling such subsets from data and then selecting the best instance \cite{xiao2024latent}. Consequently, these methods are only suitable for classical models that satisfy the minimal subset assumption. For example, a line and a circle can be determined by minimal subsets of 2 and 3 points, respectively.

\begin{figure}[!t]
\centering
\subfloat{\includegraphics[width=0.2\linewidth, valign=c]{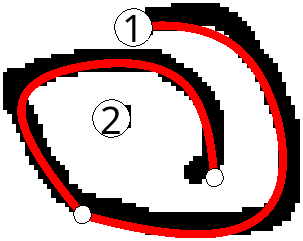}}
\hfil
\subfloat{\includegraphics[width=0.2\linewidth, valign=c]{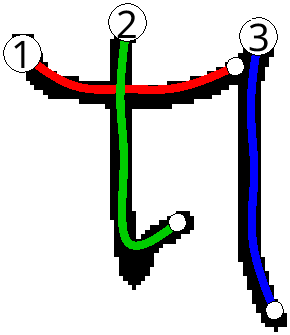}}
\hfil
\subfloat{\includegraphics[width=0.2\linewidth, valign=c]{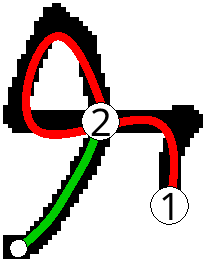}}
\hfil
\subfloat{\includegraphics[width=0.2\linewidth, valign=c]{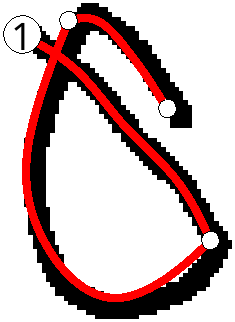}}
\caption{Four examples of non-classical models. Each example is a procedural character model presented in \cite{lake2015human}. A model consists of several strokes, which are represented by colored curves. The circled numbers indicate the starting positions of the strokes. A stroke may have sub-strokes, which are marked by white dots. }
\label{fig:character}
\end{figure}

However, for many models, identifying their minimal subsets is difficult or even impossible. We refer to these as non-classical models. For example, determining the minimal subset of a procedural character (Fig. \ref{fig:character}) or an Euler spiral \cite{zhang20193d} is non-trivial. Therefore, RANSAC-based methods are not suitable for such models.

Only a few robust fitting methods have been developed for non-classical models. The method proposed in \cite{zhang2019robust} formulates the non-classical model fitting problem as an optimization task and employs the cuckoo search (CS) algorithm \cite{yang2010engineering} as the optimizer. The objective function is defined as a robust estimator based on model-to-data error. Compared with the commonly used data-to-model error, which requires an inconvenient predefined threshold to distinguish outliers, model-to-data error can achieve robustness without such a threshold. Although this method has demonstrated robustness in a number of experiments, it suffers from the following limitation.

The method is effective at finding a single model instance but performs poorly when identifying multiple instances. In many cases, the data contain more than one model instance. To detect multiple instances, an estimator must properly handle the overlaps between different instances. However, the estimator proposed in \cite{zhang2019robust} uses the measure of model instances to regularize the model-to-data error. As a result, the overlapping regions of different instances are counted multiple times, whereas they should be seen as the same region and counted only once. In other words, the estimator cannot effectively address multi-instance model fitting.

To overcome the limitation, we propose a new robust estimator that avoids the overlapping issue. Instead of using the model measure as a regularizer, our estimator uses the number of nearest data points of model instances as the regularizer. Because overlapping regions share the same nearest data points, double counting is avoided.

In summary, our main contribution is as follows:
\begin{itemize}
\item We propose a novel estimator that (1) does not require a predefined error threshold to handle outliers, and (2) avoids the overlapping issue. With this estimator, robust fitting for multiple instances of non-classical models can be performed more effectively.
\end{itemize}

The remainder of this paper is organized as follows. Section \ref{sec:related-work} reviews relevant literature. Section \ref{sec:method} describes the proposed method. Section \ref{sec:experiments} presents the experimental results. Finally, Section \ref{sec:conclusions} concludes the paper.

\section{Related Work}\label{sec:related-work}
Model fitting is a broad research area. To fit a function-like model, such as a polynomial function or a convolutional neural network, one typically employs an estimator (e.g., mean squared error) that is differentiable with respect to the model parameters and applies gradient descent for optimization. However, for models that cannot be explicitly defined as functions—such as circles or procedural characters—defining a differentiable estimator is often non-trivial, making gradient descent not directly applicable.

In recent years, a number of deep learning–based methods have been successfully applied to fit such non-function models \cite{sharma2020parsenet}. However, these methods tend to lack robustness, as they are sensitive to outliers, particularly when the data contain multiple model instances \cite{xiao2024latent}.

Non-function models can be further divided into classical and non-classical categories. As mentioned in the previous section, RANSAC-based methods, such as Latent Semantic Consensus (LSC) \cite{xiao2024latent}, can robustly fit classical models but are difficult to extend to non-classical ones.


In summary, this paper focuses on the problem of robustly fitting multiple instances of non-classical models with a moderate number of parameters. This problem is challenging, and currently, few methods address it effectively.

\section{Proposed Method}\label{sec:method}
In this section, we describe the details of the proposed method. We first formulate the problem of multiple-instance model fitting in Section \ref{sec:problem}. Then, we introduce the proposed estimator and optimizer in Sections \ref{sec:estimator} and \ref{sec:optimizer}, respectively.

\subsection{Problem Formulation}\label{sec:problem}
A $n$-dimensional point set $P$ is a subset of $\mathbb{R}^n$, i.e., $P \subset \mathbb{R}^n$. In this paper, $n$ is 2 or 3. Let $\boldsymbol\theta$ denote the parameters of a model. A fixed value of $\boldsymbol\theta$ specifies an instance of the model. A model instance, denoted as $M_{\boldsymbol\theta}$, is in fact an infinite set of points. We refer to such a point set as a model point set. For example, a two-dimensional line segment can be represented as the following infinite set of two-dimensional points $M_{\boldsymbol\theta} \subset \mathbb{R}^2$:
\begin{equation}
M_{\boldsymbol\theta} = \{ \boldsymbol{p} + t\boldsymbol{q} | t \in [0,1] \},
\end{equation}
where $\boldsymbol{p} \in \mathbb{R}^2$, $\boldsymbol{q} \in \mathbb{R}^2$, and $\boldsymbol\theta = (\boldsymbol{p},  \boldsymbol{q})$ specifies the four parameters of the line segment.

Given a finite set of data points $D \subset \mathbb{R}^n$, multiple-instance model fitting can be formulated as the following optimization problem: 
\begin{equation}\label{eq:objective}
({\boldsymbol{\theta }_1^*},{\boldsymbol{\theta }_2^*}, \cdots ,{\boldsymbol{\theta }_{k_{\max}}^*}) = \mathop {\arg \max }\limits_{({\boldsymbol{\theta }_1},{\boldsymbol{\theta }_2}, \cdots ,{\boldsymbol{\theta }_{k_{\max}}})}  \xi( \bigcup _{k = 1}^{k_{\max}}{M_{{\boldsymbol{\theta }_k}}},D),
\end{equation}
where $k_{\max}$ denotes the number of model instances, $\xi( \cdot , \cdot )$ is an estimator that evaluates the fitness of the union of the $k_{\max}$ model instances with respect to the data $D$, and $\boldsymbol{\theta}_k^*$ denotes the parameters of the best fitted $k$-th model instance.

\subsection{Fitness Estimator}\label{sec:estimator}
We now define the estimator $\xi(\cdot , \cdot)$ used in Eq.~\eqref{eq:objective}. An estimator usually consists of two parts: an error term and a regularizer. There are two types of errors: data-to-model error and model-to-data error. As discussed in \cite{zhang2019robust}, model-to-data error is intrinsically robust to outliers, whereas data-to-model error is not. Therefore, in this paper, we adopt model-to-data error in our estimator. However, since using model-to-data error alone may result in incomplete fitting, a regularizer should be introduced to regularize the model-to-data error. It is worth noting that if data-to-model error is used instead, a regularizer is also necessary, as using data-to-model error alone may lead to overfitting.

Let $M$ be a model point set and $D$ a data point set. The model-to-data error can be defined as \cite{zhang2019robust}:
\begin{equation}\label{eq:error}
s(M, D)=\mathop {\max }\limits_{\boldsymbol{m} \in M} {\mathop {\min }\limits_{\boldsymbol{d} \in D} } \left\| {\boldsymbol{m} - \boldsymbol{d}} \right\| ,
\end{equation}
where $\left\|\cdot\right\|$ denotes the Euclidean norm.

The Mean Measure (MM) estimator proposed in \cite{zhang2019robust} uses the measure of a model to regularize the model-to-data error. In brief, the MM estimator favors models with smaller model-to-data error and larger measure. Note that the measure of a line segment is its length, whereas the measure of a surface is its area. The measure of a model can effectively serve as a regularizer for single-instance model fitting, but may fail to provide proper regularization for multiple-instance model fitting due to the overlapping issue.

\begin{figure}[!t]
\centering
\includegraphics[width=1.0\linewidth]{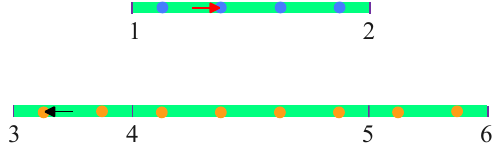}
\caption{An illustration of the overlapping issue. The blue and yellow dots represent data points. The green strips represent line segments (i.e., model instances). Let $M$ and $|M|$ denote a line segment and its length, respectively. In the figure, there are line segments $M_{12}$, $M_{34}$, $M_{36}$, and others. $|M_{12}|$ equals $|M_{45}|$. $|M_{34}|$ equals $|M_{56}|$. $|M_{34}|$ is also half of $|M_{12}|$. The model-to-data errors of each line segment are equal. For example, the $M_{12}$-to-data and $M_{34}$-to-data errors can be represented by the red arrow and the black arrow, respectively.}
\label{fig:overlap}
\end{figure}

Fig.~\ref{fig:overlap} illustrates the overlapping issue. The data point set $D$ contains 12 points, including 4 blue points and 8 yellow points. The union of line segments $M_{12} \cup M_{36}$ best fits the data, whereas the alternative union $M_{35} \cup M_{46}$ provides an incomplete fit. Therefore, the fitness of $M_{35} \cup M_{46}$ should be smaller than that of $M_{12} \cup M_{36}$.

According to Eq.~\eqref{eq:error}, the model-to-data errors of $M_{12} \cup M_{36}$ and $M_{35} \cup M_{46}$ are equal:
\begin{equation}
s(M_{12} \cup M_{36}, D) = s(M_{35} \cup M_{46}, D).
\end{equation}
Consequently, the regularizer determines which of the two unions has the larger fitness.

It can be seen that $M_{12}$ and $M_{36}$ have no overlap, whereas $M_{35}$ and $M_{46}$ overlap at $M_{45}$. If the overlap is not checked, $|M_{45}|$ is double counted and $|M_{35} \cup M_{46}|$ is incorrectly calculated as $|M_{35}|+|M_{46}|$, which equals $|M_{12} \cup M_{36}|$. Therefore, for the MM estimator \cite{zhang2019robust} that uses the measure of a model as a regularizer, the incomplete fitting model has the same fitness as the best-fit model. In other words, maximizing the estimator does not guarantee finding the best-fit model.

In summary, if the overlap is not checked, the measure of a model may not effectively serve as a regularizer for multiple-instance model fitting. Conversely, if the overlap is checked, then $M_{35} \cup M_{46}$ can be correctly treated as $M_{36}$, such that $|M_{35} \cup M_{46}|=|M_{36}|<|M_{12} \cup M_{36}|$. That is, once the overlap is checked, the measure of a model can also effectively serve as a regularizer for multiple-instance model fitting.

However, checking overlaps between model instances can be time-consuming, especially for non-classical models. Therefore, in this paper, we propose a novel estimator, the Nearest data Points Regularized model-to-data Error (NPRE). 

Let $|Q|$ denote the number of points in a finite point set $Q$. Given a model point set $M$ and a data point set $D$, the NPRE estimator is defined as:
\begin{equation}\label{eq:npre}
\xi(M,D) =  \left(\frac{\lvert A(M,D) \rvert} { \lvert D \rvert}\right)^\lambda  \frac{\delta_D } { s(M,D)},
\end{equation}
where $A(\cdot,\cdot)$ denotes the set of nearest data points of $M$, $\delta_D$ denotes the Euclidean distance between the closest pair of points in $D$ \cite{daescu2020two},  $s(\cdot,\cdot)$ can be the model-to-data error defined in Eq.~\eqref{eq:error}, and $\lambda>0$ is a hyperparameter to tune the weight of $|A(\cdot,\cdot)|$ and $s(\cdot,\cdot)$. 

Specifically, $A(\cdot,\cdot)$ is defined as:
\begin{equation}
A(M,D) = \left\{ \mathop{\arg\min}\limits_{\boldsymbol{d} \in D} 
\| \boldsymbol{m} - \boldsymbol{d} \| \;\middle|\; \boldsymbol{m} \in M \right\},
\end{equation}
and $\delta_D$ is defined as \cite{daescu2020two}\cite{zhang2019robust}:
\begin{equation}\label{eq:data_resolution}
\delta_D= \mathop {\min }\limits_{\boldsymbol{p} \in D} \mathop {\min }\limits_{\boldsymbol{q} \in D \setminus \{ \boldsymbol{p} \}  } \left\| {\boldsymbol{p} - \boldsymbol{q}} \right\|.
\end{equation}
It can be seen that, for a given set of data points $D$, the values of $|D|$ and $\delta_D$ are constant; they serve to ensure that the NPRE value is neither excessively large nor excessively small.

With the nearest data points as a regularizer, NPRE can avoid the overlapping issue, because overlapping regions of different model instances share the same nearest data points. For example, as shown in Fig.~\ref{fig:overlap}, $\lvert A(M_{12} \cup M_{36},D) \rvert = 12$, since $A(M_{12} \cup M_{36},D)$ consists of both the blue and yellow points. In contrast, $\lvert A(M_{35} \cup M_{46},D) \rvert = 8$, since $A(M_{35} \cup M_{46},D)$ consists only of the yellow points. Therefore, NPRE ensures that an incomplete fitting model has a smaller fitness than the best-fit model.

In Eq. \eqref{eq:error}, the model-to-data error is defined as the maximum distance from model $M$ to data $D$. It can also be defined as the average distance \cite{zhang2019robust}:
\begin{equation}\label{eq:average-error}
s(M, D)=\frac{1}{\lvert M^\delta \rvert} {\mathop {\sum }\limits_{\boldsymbol{m} \in M^\delta}} {\mathop {\min }\limits_{\boldsymbol{d} \in D} } \left\| {\boldsymbol{m} - \boldsymbol{d}} \right\| ,
\end{equation}
where $M^\delta$ is a finite point set uniformly sampled from $M$ at resolution $\delta$, and $\left| M^\delta \right|$ is the number of points in $M^\delta$. Note that $\delta$ is a fixed parameter that does not need to be tuned, as it depends only on the data resolution (i.e., $\delta_D$ in Eq. \eqref{eq:data_resolution}) \cite{zhang2019robust}. We use the maximum distance (Eq. \eqref{eq:error}) only for illustration (Fig. \ref{fig:overlap}). In the experiments, we use the average distance (Eq. \eqref{eq:average-error}) to compute the model-to-data error for NPRE (Eq. \eqref{eq:npre}), since the average distance makes the optimization smoother than the maximum distance.

\subsection{Optimizer}\label{sec:optimizer}
Our method sequentially fits multiple model instances one by one. Specifically, each instance is fitted using the cuckoo search (CS) algorithm \cite{yang2010engineering}. The CS algorithm works iteratively and consists of the following four steps: initialization, perturbation, selection, and recombination \cite{camacho2022analysis}.

Step 1: initialization. Let the population size of cuckoos (i.e., the number of solutions) be $n_p$. To fit $k$-th model instance, randomly initialize $n_p$ solutions ($cuckoos$) $\boldsymbol{\theta}_{k,i}^j$:
\begin{equation}
\label{eq:initial}
\theta_{k,i=0}^{j, m} = \mathcal{U}[\theta_{\text{min}}^m, \theta_{\text{max}}^m],~~ \text{for}~j = 1, \cdots, n_p ~ \text{and} ~ m = 1, \cdots, n_{\boldsymbol{\theta}},
\end{equation}
where $i$ is the iteration count, $\mathcal{U}$ is a random uniform distribution, $n_{\boldsymbol{\theta}}$ is the number of model parameters $\boldsymbol{\theta}$ (see Eq. \eqref{eq:objective}), and $\theta_{\text{min}}^m$ and $\theta_{\text{max}}^m$ are the lower and upper bounds of the $m$-th parameter of $\boldsymbol{\theta}$. 

Step 2: perturbation. This step is also called Lévy flight, and its implementation details are as follows. Perturb all $n_p$ solution $\boldsymbol{\theta}_{k,i}^j$ to obtain new solutions (i.e., eggs) by adding a random vector, as shown below:
\begin{equation}
\label{eq:mutate}
\boldsymbol{\theta}_{k,i}^{j'} = \boldsymbol{\theta}_{k,i}^{j}+\boldsymbol{\alpha} \odot \boldsymbol{w}_{k,i}^{j}, ~\text{for}~j = 1 , \cdots, n_p,
\end{equation}
where $\boldsymbol{\theta}_{k,i}^{j'}$ is the perturbed solution, $\boldsymbol{w}_{k,i}^{j}$ is a random vector whose components are sampled from the standard Gaussian distribution, $\odot$ denotes element-wise multiplications, and $\boldsymbol{\alpha}$ is a vector with the same size as $\boldsymbol{\theta}$ that controls the magnitude of the perturbation. Specifically,
\begin{equation}
\boldsymbol{\alpha} = 0.01 \eta (\boldsymbol{\theta}_{k,i}^{j} - \boldsymbol{\theta}_{k,i}^{j*}),
\end{equation}
where
\begin{equation}
\eta  = \frac{u}{{{{\left| v \right|}^{\frac{1}{\beta }}}}},
\end{equation}
with $\beta=1.5$, $v$ being a random variable sampled from the standard Gaussian distribution, i.e., $v \sim \mathcal{N}(0,1)$, and $u$ a random variable sampled from a Gaussian distribution $\mathcal{N}(0,\sigma _u^2)$, where 
\begin{equation}
{\sigma _u} = {\left( {\frac{{\Gamma (1 + \beta )\sin (\frac{{\pi \beta }}{2})}}{{\Gamma (\frac{{1 + \beta }}{2}\beta {2^{\frac{{\beta  - 1}}{2}}})}}} \right)^{\frac{1}{\beta }}},
\end{equation}
where $\Gamma(\cdot) $ is the gamma function \cite{cody2006overview}.

Step 3: selection. 
The objective function $f(\cdot)$  for fitting $k$-th model instance is defined as:
\begin{equation}
\label{eq:multi-instance-objective}
f\left(\boldsymbol{\theta } \right) = \xi\left( M_{k-1} \cup M_{\boldsymbol{\theta }},D \right), 
\end{equation}
where $\xi(\cdot,\cdot)$ is the NPRE estimator defined in Eq. \eqref{eq:npre}, $D$ is the given data point set, and $M_{k-1}$ denotes the union of previous fitted $k-1$ instances:
\begin{equation}
M_{k-1}=\bigcup _{t = 1}^{k-1}{M_{{\boldsymbol{\theta }_{t,*}}}} \, ,
\end{equation}
where $\boldsymbol{\theta }_{t,*}$ denotes the parameters of the best fitted $t$-th model instance.

Compare each pair $({\boldsymbol{\theta }}_{k,i}^j, \boldsymbol{\theta }_{k,i}^{j'})$ based on the objective function $f(\cdot)$ defined in Eq. \eqref{eq:multi-instance-objective} and select the one with the higher quality. This is formally expressed as follows:
\begin{equation}
\label{eq:selection}
{\boldsymbol{\theta }}_{k,i'}^j = \left\{ \begin{array}{l}
{\boldsymbol{\theta }}_{k,i}^{j'},{\rm{~~if }}~f(\boldsymbol{\theta }_{k,i}^{j'}){\rm{~is~better~than~}}f({\boldsymbol{\theta }}_{k,i}^j),\\
{\boldsymbol{\theta }}_{k,i}^j,{\rm{~~otherwise.}}
\end{array} \right. 
\end{equation}

Step 4: recombination. Let the discovery rate of alien eggs be denoted as $p_a$. In this paper, $p_a$ is set to 0.25. With probability $1-p_a$, apply recombination to the $m$-th component of the vector ${\boldsymbol{\theta }}_{k,i'}^j$ using two randomly selected solutions $\boldsymbol{\theta}_{k,i'}^{h^j} \in H_i$ and $\boldsymbol{\theta}_{k,i'}^{g^j} \in G_i$, where sets $H_i$ and $G_i$ contain eash a copy of the population after executing step 3: selection, i.e., a copy of ${\boldsymbol{\theta }}_{k,i'}^j$, for $j=1,\cdots,n_p$. Recombination is calculated as follows. $\forall m,\forall j$:
\begin{equation}
\label{eq:recombination}
\theta _{k,i + 1}^{j,m} = \left\{ \begin{array}{l}
\theta _{k,i'}^{j,m} + \mathcal{U}[0,1] \cdot (\theta _{k,i'}^{{h^j},m} - \theta _{k,i'}^{{g^j},m}),{\rm{~~if~}}\mathcal{U}[0,1] \ge {p_a},\\
\theta _{k,i'}^{j,m},{\rm{~~otherwise.}}
\end{array} \right.
\end{equation}
Solutions $\boldsymbol{\theta}_{k,i'}^{h^j}$ and $\boldsymbol{\theta}_{k,i'}^{g^j}$ are selected from sets $H_i$ and $G_i$ without replacement, that is, each solution is used once as $\boldsymbol{\theta}_{k,i'}^{h}$ and once as $\boldsymbol{\theta}_{k,i'}^{g}$. After finishing the recombination, solutions are evaluated once again.

The pseudo-code for fitting the $k$-th model instance is presented in Algorithm \ref{alg:single}, where the best solution ${\boldsymbol\theta}_{k,*}$ is returned as the parameters of the best fitted $k$-th model instance. Note that, the best $cuckoo$ or $egg$ in each iteration is denoted as $\boldsymbol{\theta}_{k, i}^{t^*}$, where
\begin{equation}
\label{eq:best}
t^* = {\mathop{\arg\max}\limits_{1 \leq t \leq n_p}  \ f\left(\boldsymbol{\theta }_{k,i}^{t} \right)}, 
\end{equation}
where $f(\cdot)$ is the objective function defined in Eq. \eqref{eq:multi-instance-objective}, and $\boldsymbol{\theta }_{k,i}^{t}$ represents the $t$-th $cuckoo$ or $egg$ in the $i$-th iteration.

\begin{algorithm}
\small
\caption{Fitting for the $k$-th model instance }
\label{alg:single}
\begin{algorithmic}[1]
\State \textbf{input}: the number of cuckoos $n_p$, the maximum number of iterations $i_{\max}$
\State $i \gets 0$
\State initialize the best fitness of the $k$-th instance: $f_{k,*} \gets 0$
\State initialize $n_p$ $cuckoos$ (solutions) \Comment{Eq. \eqref{eq:initial}}
\State \text{evaluate the $n_p$ $cuckoos$} \Comment{Eq. \eqref{eq:multi-instance-objective}}
\State \text{obtain the best $cuckoo$ $\boldsymbol{\theta}_{k, i}^{t^*}$} \Comment{Eq. \eqref{eq:best}}
\If{$f(\boldsymbol{\theta}_{k, i}^{t^*}) > f_{k,*}$ }
\State $f_{k,*} \gets f(\boldsymbol{\theta}_{k, i}^{t^*})$
\State ${\boldsymbol\theta}_{k,*} \gets \boldsymbol{\theta}_{k, i}^{t^*}$
\EndIf
\While{$i < i_{\max}$}
\State mutate the $n_p$ $cuckoos$ to get $n_p$ $eggs$ \Comment{Eq. \eqref{eq:mutate}}
\State evaluate the $n_p$ $eggs$ \Comment{Eq. \eqref{eq:multi-instance-objective}}
\State \text{obtain the best $egg$ $\boldsymbol{\theta}_{k, i}^{t^*}$} \Comment{Eq. \eqref{eq:best}}
\If{$f(\boldsymbol{\theta}_{k, i}^{t^*}) > f_{k,*}$ }
\State $f_{k,*} \gets f(\boldsymbol{\theta}_{k, i}^{t^*})$
\State ${\boldsymbol\theta}_{k,*} \gets \boldsymbol{\theta}_{k, i}^{t^*}$
\EndIf
\State select $n_p$ solutions from the $cuckoos$ and $eggs$ \Comment{Eq. \eqref{eq:selection}}
\State recombine the $n_p$ selected solutions \Comment{Eq. \eqref{eq:recombination}}
\State use the $n_p$ recombined solutions as new $cuckoos$
\State \text{evaluate the $n_p$ $cuckoos$} \Comment{Eq. \eqref{eq:multi-instance-objective}}
\State \text{obtain the best $cuckoo$ $\boldsymbol{\theta}_{k, i}^{t^*}$} \Comment{Eq. \eqref{eq:best}}
\If{$f(\boldsymbol{\theta}_{k, i}^{t^*}) > f_{k,*}$ }
\State $f_{k,*} \gets f(\boldsymbol{\theta}_{k, i}^{t^*})$
\State ${\boldsymbol\theta}_{k,*} \gets \boldsymbol{\theta}_{k, i}^{t^*}$
\EndIf
\State $i \gets i+1$ 
\EndWhile
\State \Return the best solution ${\boldsymbol\theta}_{k,*}$ as fitted $k$-th model instance
\end{algorithmic}
\end{algorithm}

\section{Experiments}\label{sec:experiments}
\subsection{Line Fitting}
\begin{figure}[!t]
\centering
\subfloat{\includegraphics[width=0.23\linewidth]{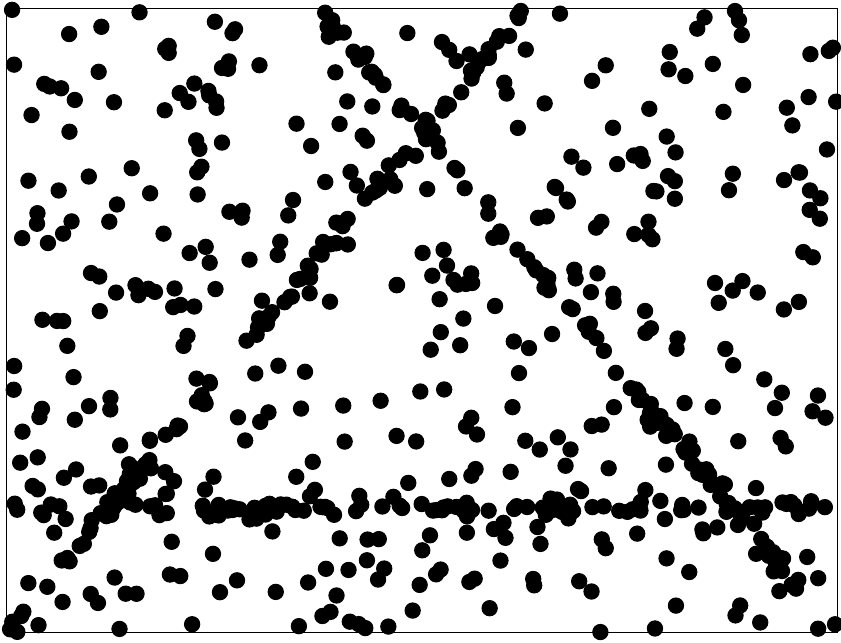}}
\hfil
\subfloat{\includegraphics[width=0.23\linewidth]{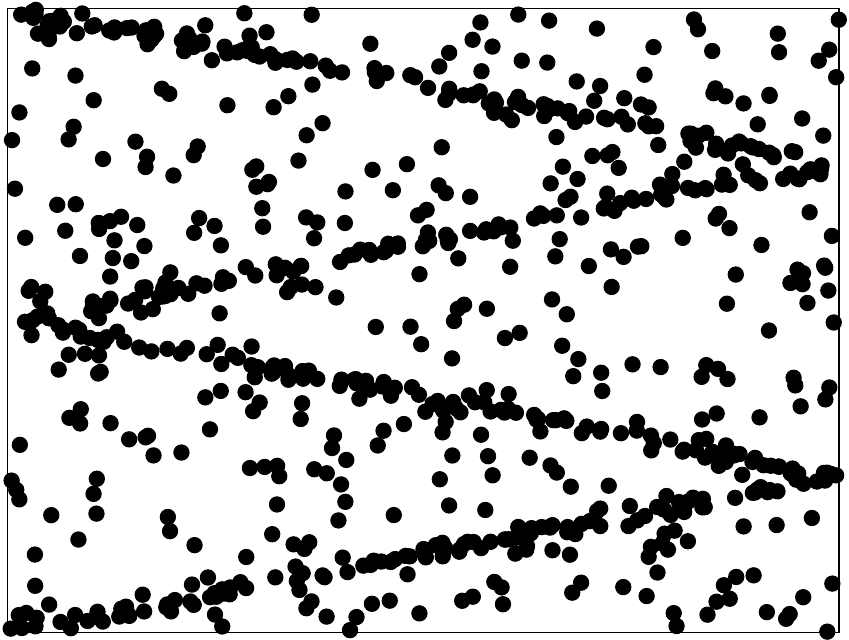}}
\hfil
\subfloat{\includegraphics[width=0.23\linewidth]{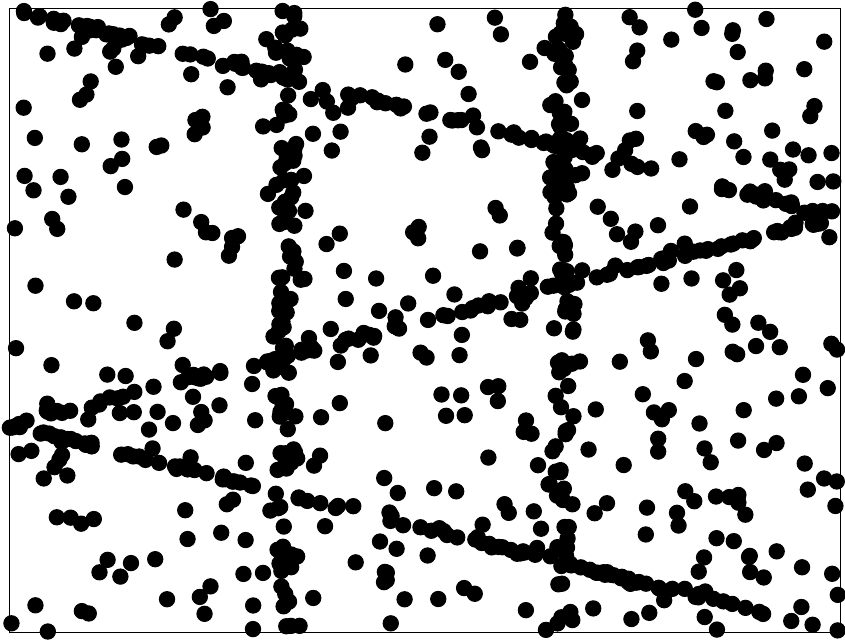}}
\hfil
\subfloat{\includegraphics[width=0.23\linewidth]{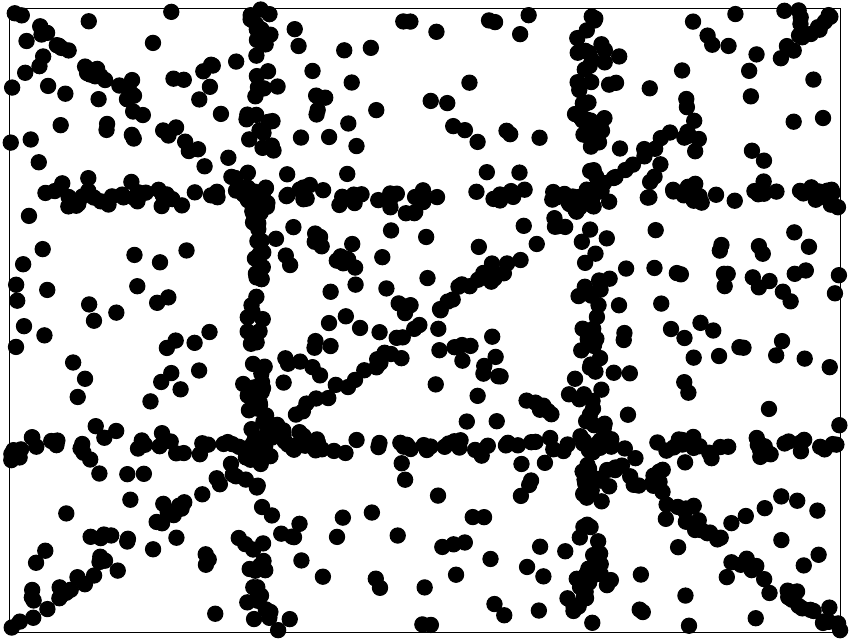}}
\\

\subfloat{\includegraphics[width=0.23\linewidth]{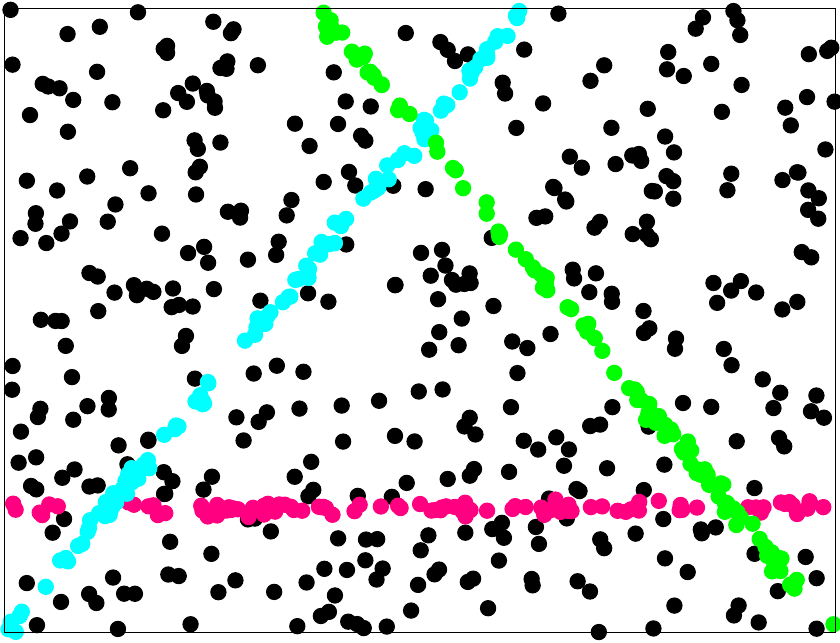}}
\hfil
\subfloat{\includegraphics[width=0.23\linewidth]{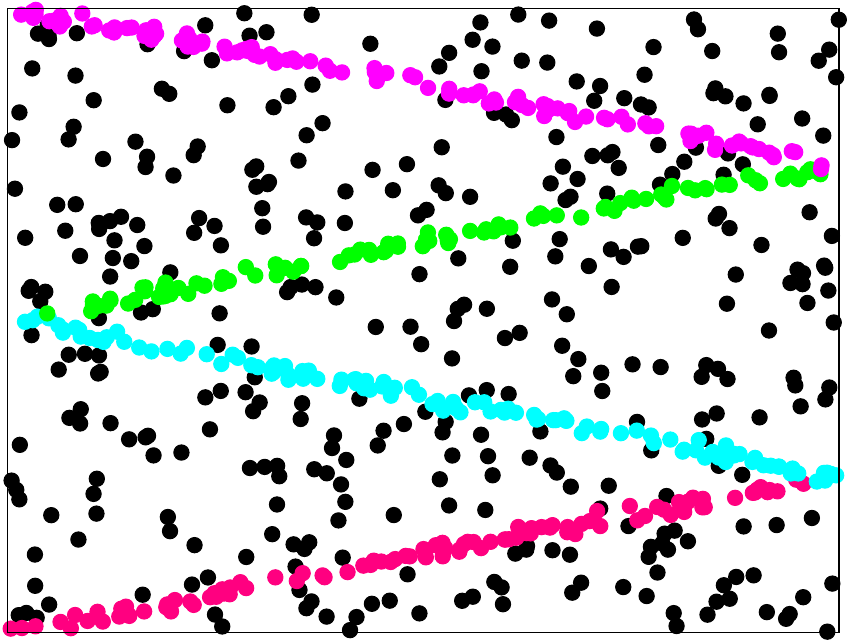}}
\hfil
\subfloat{\includegraphics[width=0.23\linewidth]{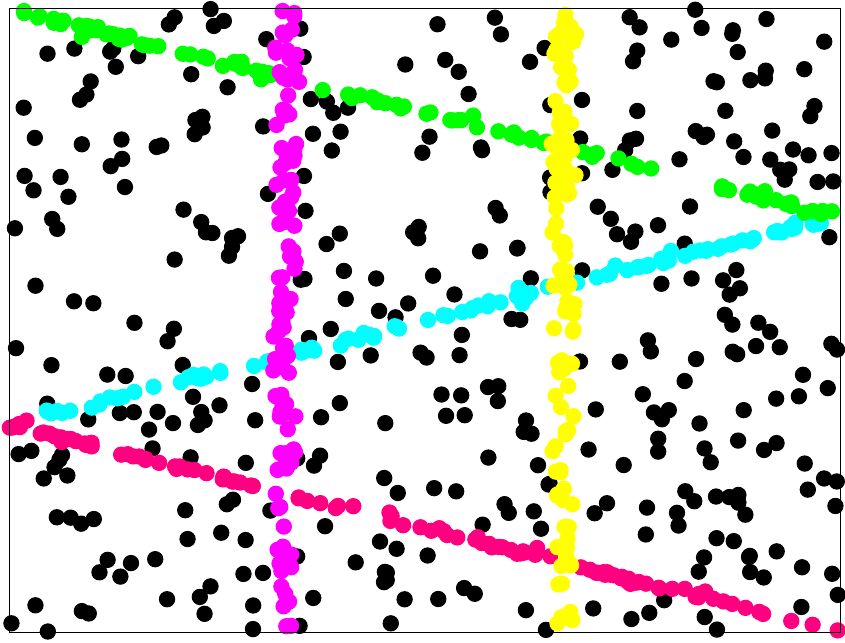}}
\hfil
\subfloat{\includegraphics[width=0.23\linewidth]{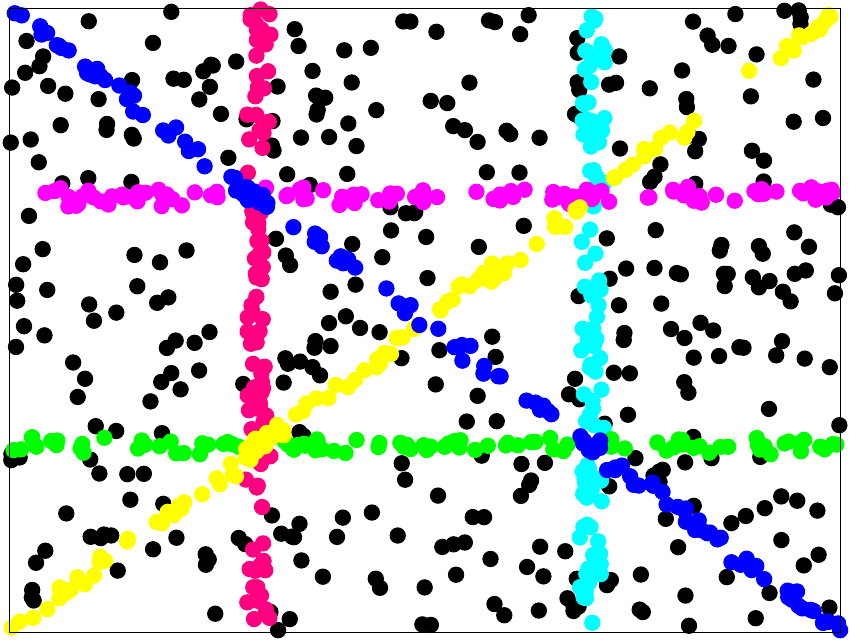}}
\\

\subfloat{\includegraphics[width=0.23\linewidth, clip=true, trim=30mm 0mm 30mm 10mm]{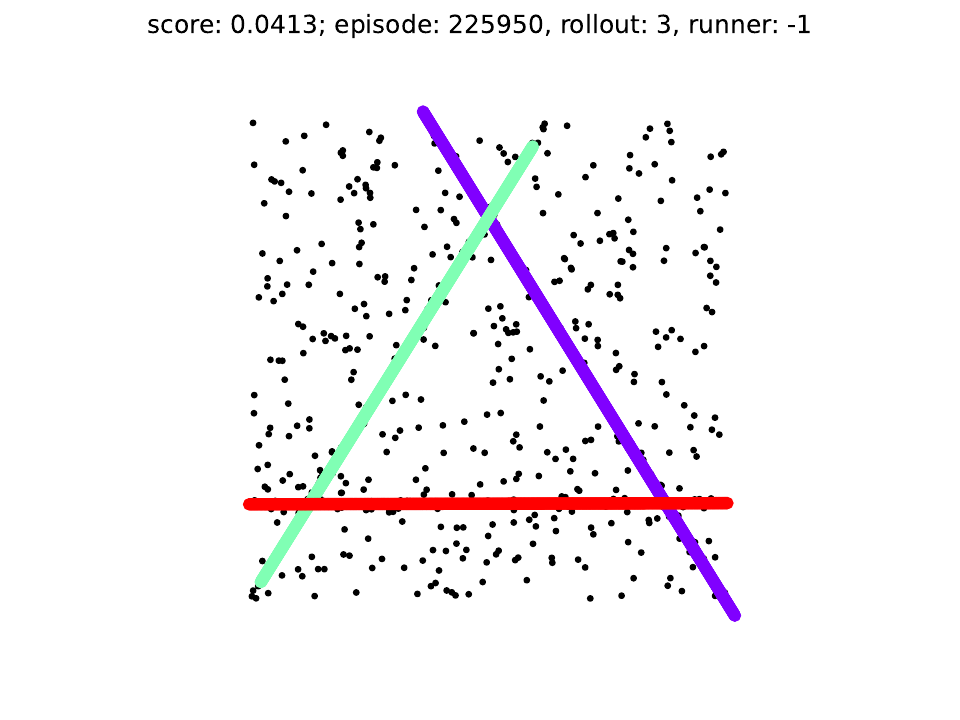}}
\hfil
\subfloat{\includegraphics[width=0.23\linewidth, clip=true, trim=30mm 0mm 30mm 10mm]{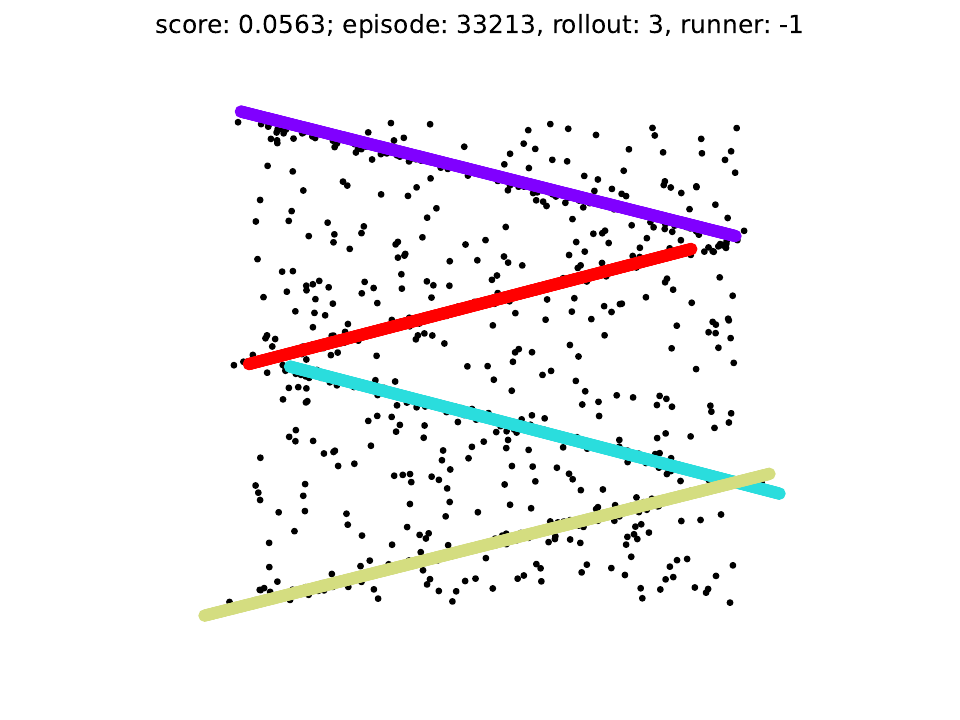}}
\hfil
\subfloat{\includegraphics[width=0.23\linewidth, clip=true, trim=30mm 0mm 30mm 10mm]{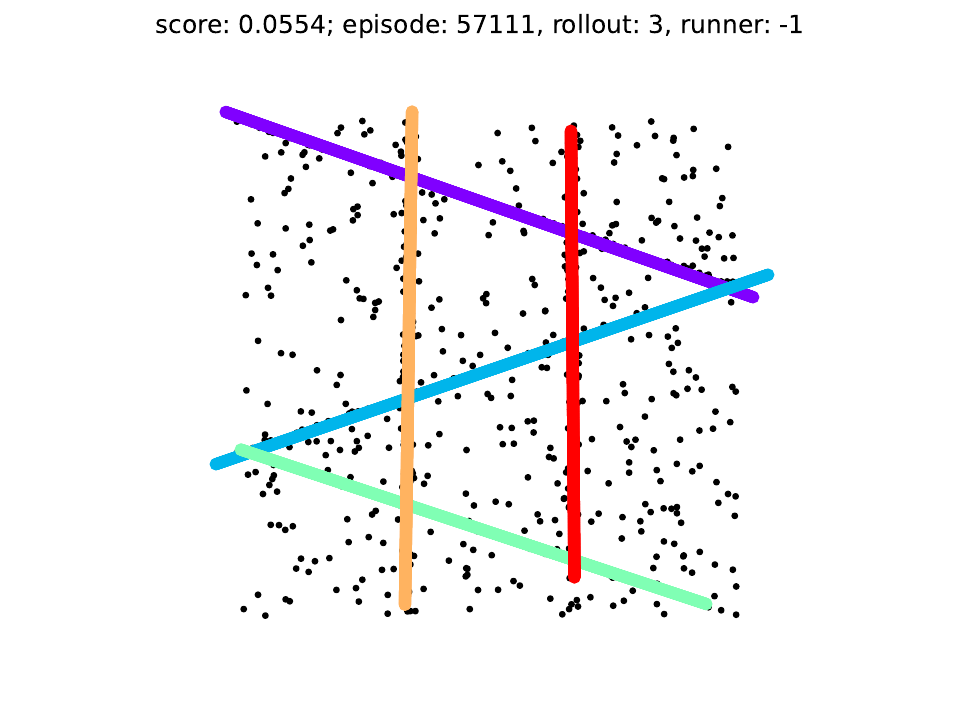}}
\hfil
\subfloat{\includegraphics[width=0.23\linewidth, clip=true, trim=30mm 0mm 30mm 10mm]{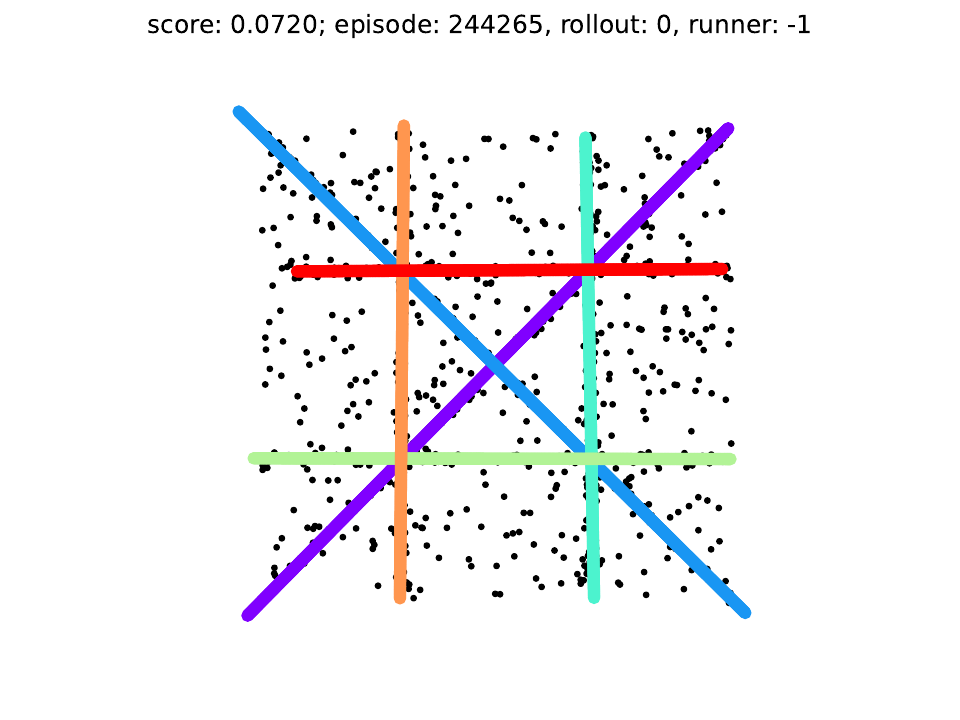}}
\caption{Fitting on the noisy line data. The first row shows the data. The second row shows the results obtained by the state-of-the-art classical model fitting method \cite{xiao2024latent}. The third row shows the results obtained by our method.}
\label{fig:line}
\end{figure}

Our method is designed for non-classical models, and can also be applied to classical models. Fig. \ref{fig:line} shows the fitting results on the data contaminated by severe outliers. It can be seen that the robustness of our method to outliers is comparable with the state-of-the-art robust fitting method for classical models \cite{xiao2024latent}.

\subsection{Character Fitting}

We now present experimental results of fitting non-classical models. The models are the procedural character models presented in \cite{lake2015human}. In these character models, a character is composed of several strokes, and the parameters can be categorized into character-wise and stroke-wise parameters. Character-wise parameters include the rotation, position, and scale of a character. Stroke-wise parameters include the rotation, position, scale, width, and shape of a stroke. The shape of a stroke is represented by a B-spline curve with five control points. Each control point has two variables (horizontal and vertical position). Thus, there are $5 \times 2 = 10$ variables for the shape per stroke \cite{peng2025robust}. In brief, a character with more strokes has more parameters.

\begin{figure}[!t]
\centering
\subfloat[]{\label{fig:character3:noisy}\includegraphics[width=0.45\columnwidth]{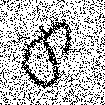}}
\hfill
\subfloat[]{\label{fig:character3:cs}\includegraphics[width=0.45\columnwidth, clip=true, trim=10mm 0mm 10mm 10mm]{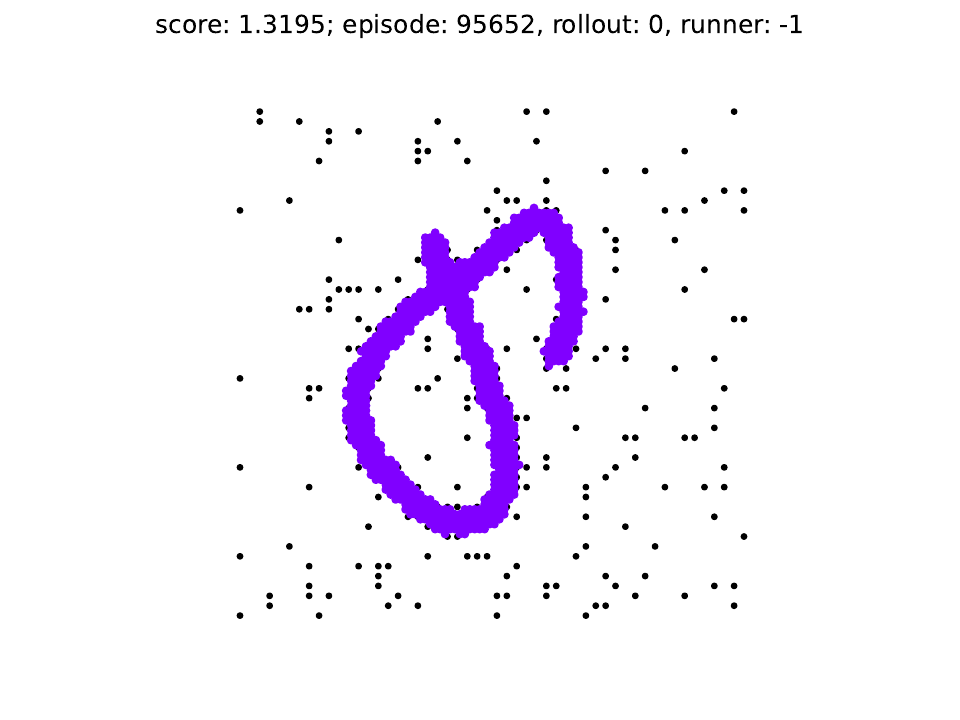}}

\caption{Fitting on character data. \protect\subref{fig:character3:noisy} Noisy image. \protect\subref{fig:character3:cs} Results of fitting the character model on \protect\subref{fig:character3:noisy}.}
\label{fig:character3}
\end{figure}

Fig. \ref{fig:character3} shows the result of fitting character model which has 31 parameters. Fig. \ref{fig:character3:noisy} shows a noisy image corrupted by salt-and-pepper noise \cite{peng2025robust}. The image is grayscale and is converted to a 2-dimensional point set by binarization \cite{peng2025robust}. The fitting is then performed on the point set. The number of fitness evaluations is 100,000. Fig. \ref{fig:character3:cs} shows the fitting result obtained by our method. It can be seen that our method has successfully reconstructed all the strokes from noisy image, demonstrating that our proposed NPRE estimator is robust against severe noise.

\subsection{Road Curve Fitting}
\begin{figure}[!t]
\centering

\subfloat[]{\label{fig:circle-parabola:data}\includegraphics[width=1\columnwidth]{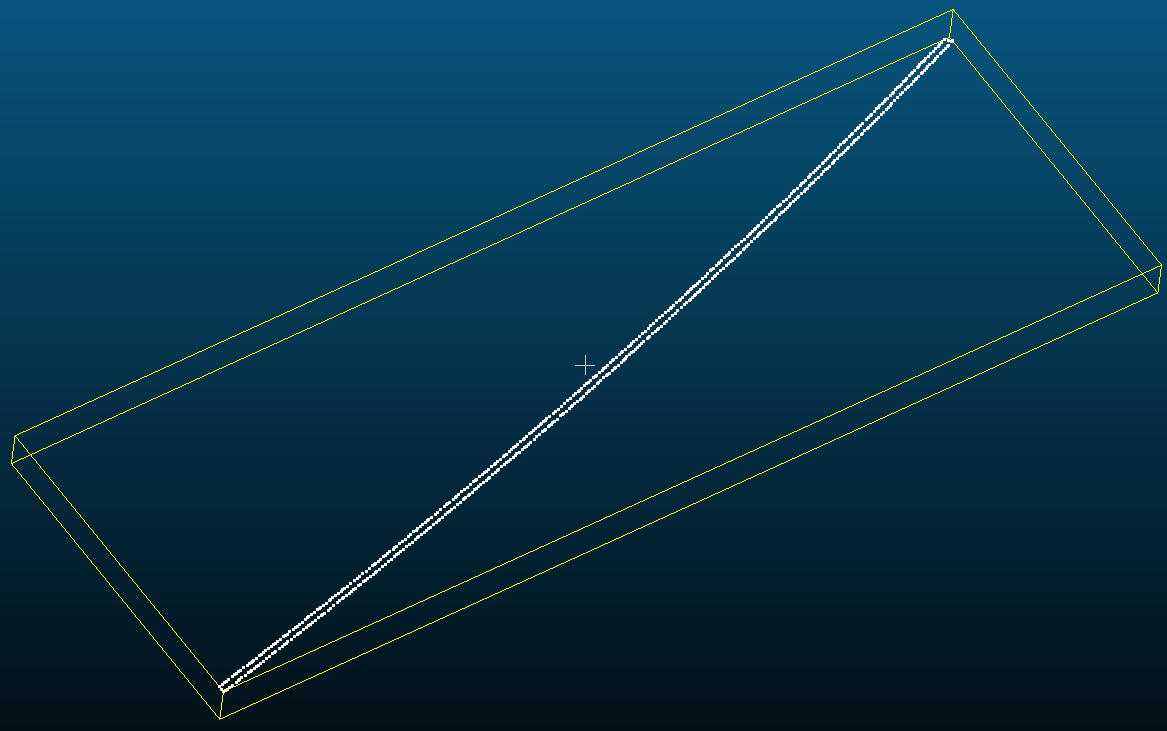}}
\hfill
\subfloat[]{\label{fig:circle-parabola:model}\includegraphics[width=1\columnwidth]{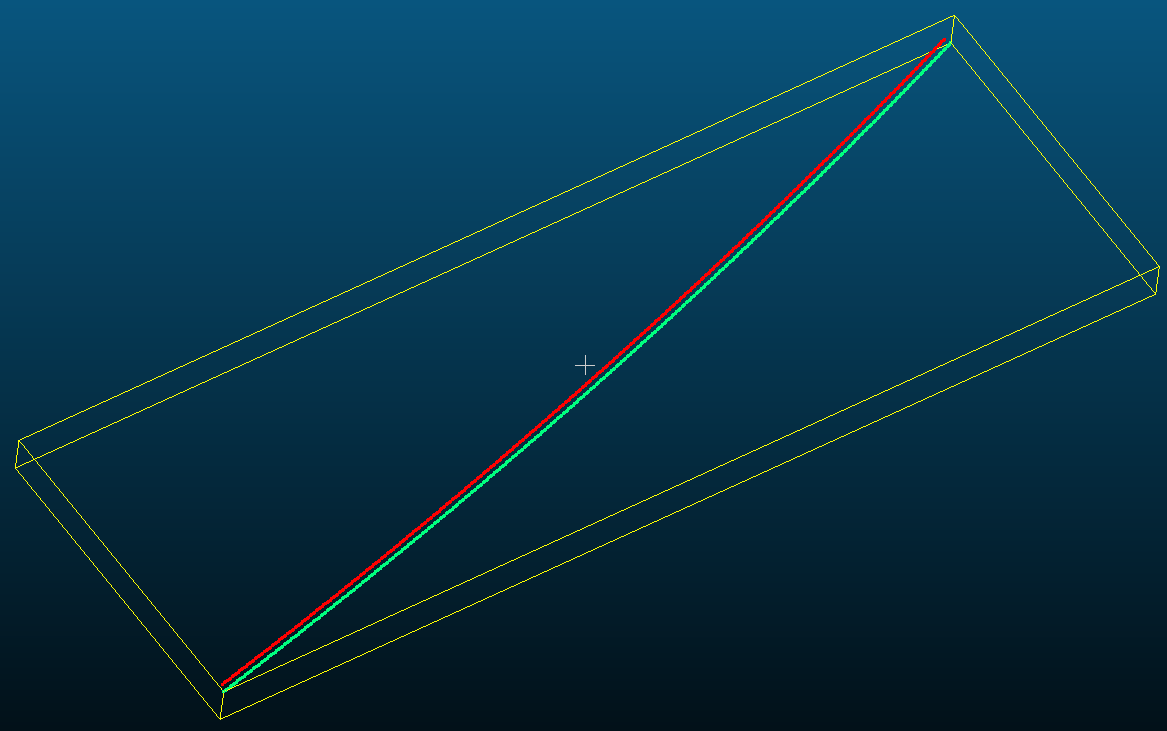}}
\hfill
\subfloat[]{\label{fig:circle-parabola:overlap}\includegraphics[width=1\columnwidth]{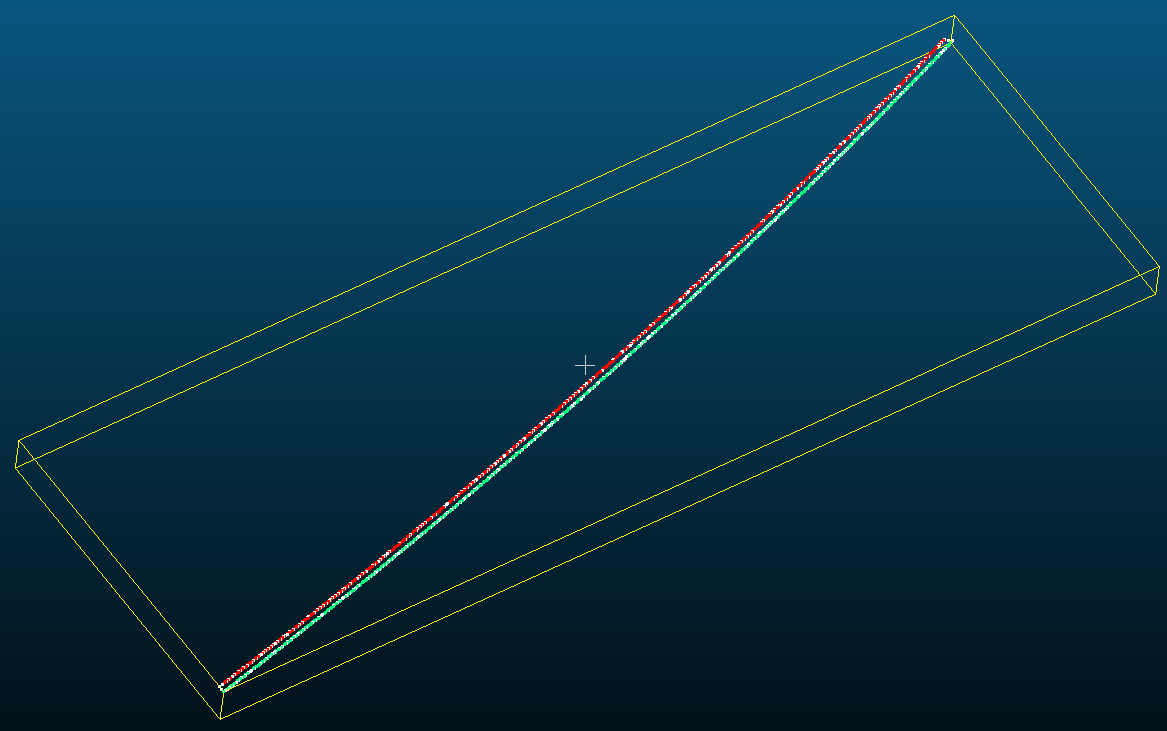}}

\caption{Fitting on synthetic road curve data \cite{zhang20193d}. \protect\subref{fig:circle-parabola:data} the synthetic 3D point cloud data (white). \protect\subref{fig:circle-parabola:model} The two road curves (red and green) reconstructed from \protect\subref{fig:circle-parabola:data} by our method. \protect\subref{fig:circle-parabola:overlap} The two reconstructed road curves (red and green) are shown together with the data (white).}
\label{fig:circle-parabola}
\end{figure}

Fig. \ref{fig:circle-parabola} and Table \ref{tab:circle-parabola} show the results of fitting 3D road curves on the synthetic data \cite{zhang20193d}. The model used to fit is the 3D highway curve model with horizontal circle and vertical parabola \cite{zhang20193d}. The model has eight parameters, including three parameters for the start location, one parameter for the curve length, one parameter for the start horizontal azimuth, one parameter for the start slope, one parameter for the horizontal radius, one parameter for the vertical curvature.

It can be seen from Fig. \ref{fig:circle-parabola} that, our method accurately reconstructs two road curves from the data, while the method proposed in \cite{zhang20193d} is only able to reconstruct one curve. The road curve model used in this experiment has two critical parameters: horizontal radius and vertical curvature. It can be seen from Table \ref{tab:circle-parabola} that the estimated values of the two critical parameters are close to the ground-truth values, especially for the parameter of vertical curvature.

\begin{figure}[!t]
\centering

\subfloat[]{\label{fig:spiral-parabola:data}\includegraphics[width=1\columnwidth]{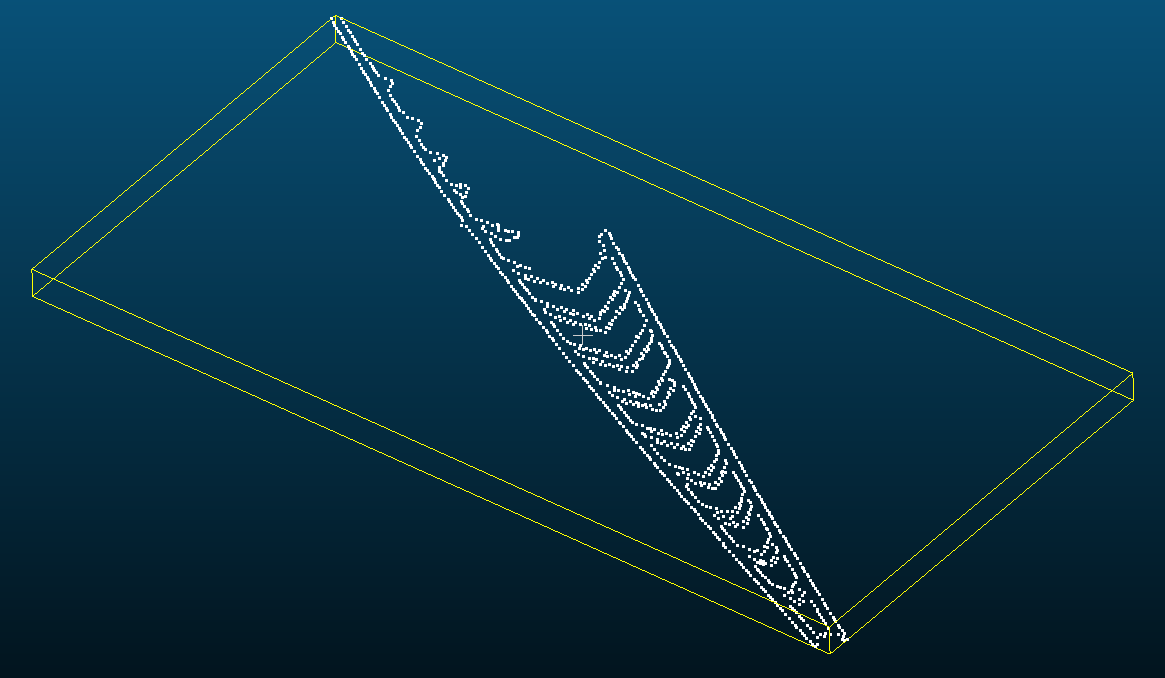}}
\hfill
\subfloat[]{\label{fig:spiral-parabola:model}\includegraphics[width=1\columnwidth]{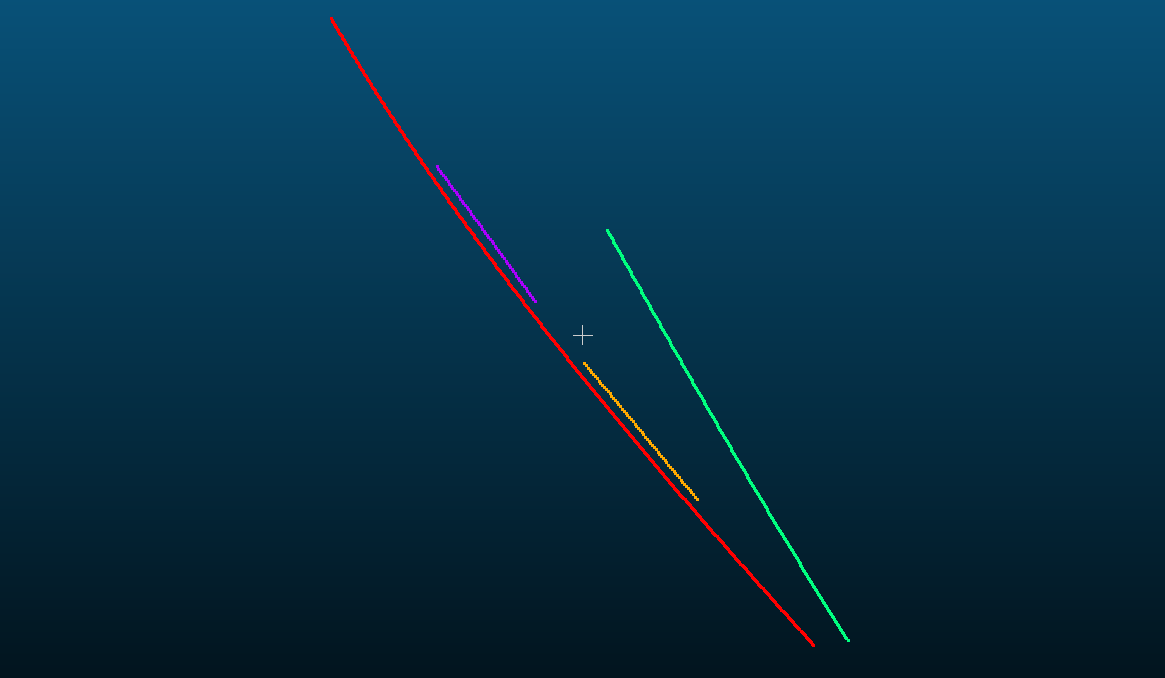}}
\hfill
\subfloat[]{\label{fig:spiral-parabola:overlap}\includegraphics[width=1\columnwidth]{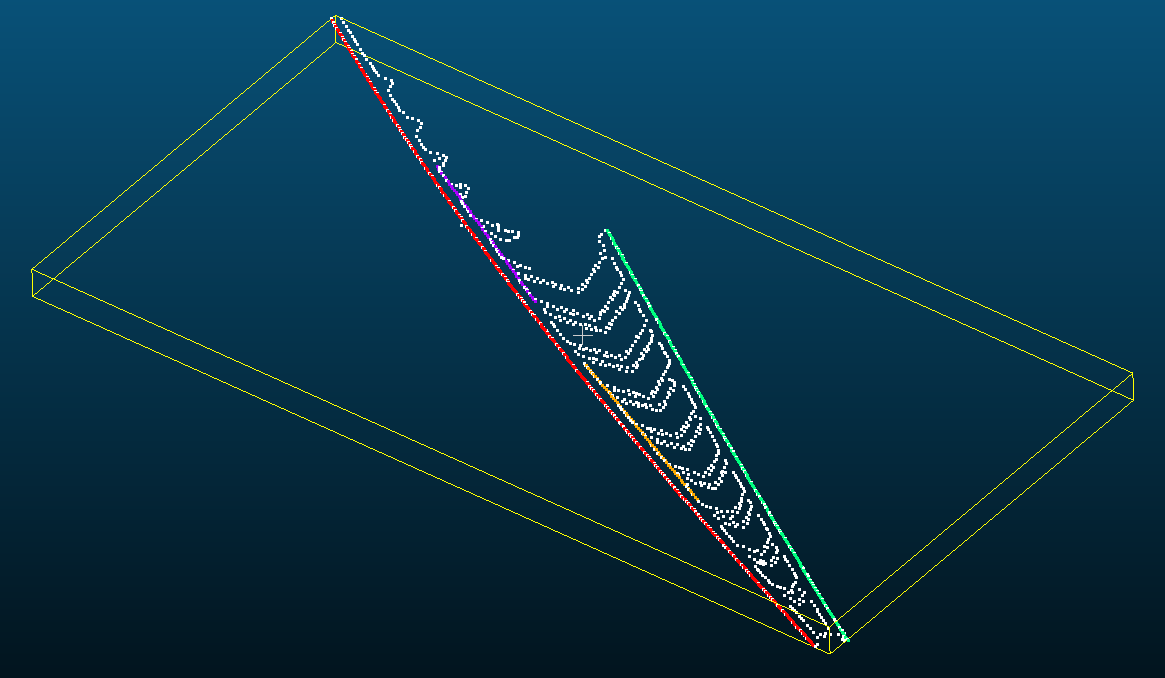}}

\caption{Fitting on the real-world laser scanning pint cloud data \cite{zhang20193d}. \protect\subref{fig:spiral-parabola:data} the 3D laser scanning point cloud data (white). \protect\subref{fig:spiral-parabola:model} The four road curves (color) reconstructed from \protect\subref{fig:spiral-parabola:data} by our method. \protect\subref{fig:spiral-parabola:overlap} The four reconstructed road curves (color) are shown together with the data (white).}
\label{fig:spiral-parabola}
\end{figure}

We also conduct experiments on real-world laser scanning pint cloud data \cite{zhang20193d}, by using 3D highway curve model with horizontal spiral and vertical parabola. As shown in Fig. \ref{fig:spiral-parabola}, our method is able to successfully reconstruct multiple highway curves from the laser scanning data.


\begin{table*}[]
\centering
\caption{The results of fitting on the synthetic road curve data (see Fig. \ref{fig:circle-parabola}).}
\begin{tabular}{|l|ll|ll|}
\hline
                     & \multicolumn{2}{l|}{Red curve}                                  & \multicolumn{2}{l|}{Green Curve}                              \\ \hline
\textbf{Parameter}            & \multicolumn{1}{l|}{\textbf{Horizontal radius}}  & \textbf{Vertical curvature}    & \multicolumn{1}{l|}{\textbf{Horizontal radius}}  & \textbf{Vertical curvature}  \\ \hline
Ground-truth         & \multicolumn{1}{l|}{349}                & 0.0005                & \multicolumn{1}{l|}{349.2}              & 0.0005              \\ \hline
Mean measure \cite{zhang20193d} & \multicolumn{1}{l|}{Not reconstructed}                   &     Not reconstructed                  & \multicolumn{1}{l|}{347.2981}           & 0.00049981          \\ \hline
Our method           & \multicolumn{1}{l|}{348.78380145810206} & 0.0004977896908984532 & \multicolumn{1}{l|}{347.43319601057755} & 0.00050003650335783 \\ \hline
\end{tabular}
\label{tab:circle-parabola}

\end{table*}

\section{Conclusions}\label{sec:conclusions}

In this paper, we presented a novel approach for the robust fitting of multiple instances of non-classical geometric models. Unlike classical models that satisfy the minimal subset assumption required by RANSAC-based methods, non-classical models like procedural characters and spirals present significant challenges due to their complex parameterizations.

The core of our contribution is the NPRE estimator. By utilizing the number of nearest data points as a regularizer, our method effectively resolves the overlapping issue where different instances sharing the same region were potentially double-counted. This estimator is intrinsically robust to outliers and does not require a manually tuned error threshold. To optimize this non-differentiable estimator, we employed the CS algorithm to fit instances sequentially. Experimental results across various datasets—including noisy lines, procedural characters, and 3D highway curves—demonstrate that our method is able to precisely reconstruct multiple instances of geometric models from noisy data.

Future work will focus on improving the computational efficiency of the optimization process to handle even more complex non-classical models with a higher number of parameters in real-time applications.

\end{document}